\documentclass[sigconf]{aamas} 
\pdfoutput=1
\renewcommand\footnotetextcopyrightpermission[1]{} 
\usepackage{balance} 
\usepackage{url}
\usepackage{graphicx}
\usepackage{orcidlink}

\title{Demonstrating Performance Benefits of Human-Swarm Teaming}

\author{William Hunt\, \orcidlink{0000-0003-4269-5050}}
\affiliation{
  \institution{University of Southampton}
  \city{Southampton}
  \country{United Kingdom}}
\email{w.hunt@soton.ac.uk}

\author{Jack Ryan\, \orcidlink{0009-0002-5215-1006}}
\affiliation{
  \institution{University of Southampton}
  \city{Southampton}
  \country{United Kingdom}}
\email{j.ryan@soton.ac.uk}

\author{Ayodeji O. Abioye\, \orcidlink{0000-0003-4637-3278}}
\affiliation{
  \institution{University of Southampton}
  \city{Southampton}
  \country{United Kingdom}}
\email{a.o.abioye@soton.ac.uk}

\author{Sarvapali D. Ramchurn\, \orcidlink{0000-0001-9686-4302}}
\affiliation{
  \institution{University of Southampton}
  \city{Southampton}
  \country{United Kingdom}}
\email{sdr1@soton.ac.uk}

\author{Mohammad D. Soorati\, \orcidlink{0000-0001-6954-1284}}
\affiliation{
  \institution{University of Southampton}
  \city{Southampton}
  \country{United Kingdom}}
\email{m.soorati@soton.ac.uk}

\begin{abstract}
Autonomous swarms of robots can bring robustness, scalability and adaptability to safety-critical tasks such as search and rescue but their application is still very limited. Using semi-autonomous swarms with human control can bring robot swarms to real-world applications. Human operators can define goals for the swarm, monitor their performance and interfere with, or overrule, the decisions and behaviour.
We present the ``Human And Robot Interactive Swarm'' simulator (HARIS) that allows multi-user interaction with a robot swarm and facilitates qualitative and quantitative user studies through simulation of robot swarms completing tasks, from package delivery to search and rescue, with varying levels of human control. In this demonstration, we showcase the simulator by using it to study the performance gain offered by maintaining a ``human-in-the-loop'' over a fully autonomous system as an example. This is illustrated in the context of search and rescue, with an autonomous allocation of resources to those in need.
\end{abstract}

         
\newcommand{\BibTeX}{\rm B\kern-.05em{\sc i\kern-.025em b}\kern-.08em\TeX}

\begin{document}

\pagestyle{fancy}
\fancyhead{}

\maketitle 

\section{Introduction}
Autonomous multi-agent systems are becoming increasingly ubiquitous in our everyday lives, from autonomous vehicles to home automation, and most pressingly in high-stakes scenarios such as disaster relief~\cite{innocente2019self, divband2021designing}. In these settings, more so than any other, the cognitive burden on human operators to control a large collective of autonomous systems over a dispersed area is unmanageable~\cite{st2019planetary}. A single human operator is unable to effectively control individual agents, designate tasks, and achieve overall mission objectives in a timely and precise manner. On the other hand, fully autonomous systems are still lacking in their ability to handle a diverse range of real-world scenarios reliably, and as such their performance is not assured, nor deemed trustworthy, for safety-critical systems.~\cite{parnell2022trustworthy, nam2019models}

It is hypothesised that the optimal solution lies at the intersection between these two approaches – in human-swarm teaming~\cite{hussein2018mixed} – and is therefore the primary goal of our proposed tool. An efficient human-swarm system requires a highly usable interface for the human operator(s) of the swarm to define tasks with priorities, allocate groups of agents to a task set, monitor the behaviour of the swarm, analyze the decisions of the system and intervene when needed~\cite{divband2021designing}.

The objective is to enable human-swarm teaming in various scenarios and allow researchers and practitioners to quickly and easily build and run user studies with a human-supervised swarm of robots. To achieve this objective, we extend a digital twin simulation platform, the Human Teaming Simulator (HutSim)~\cite{Ramchurn2015a}, that facilitates the completion of a search and rescue task, in two different modes---fully autonomous and human-teaming modes. Existing multi-agent simulators such as Gazebo~\cite{koenig2004design}, Webots~\cite{michel2004cyberbotics}, and ARGoS~\cite{pinciroli2012argos} have limited utility for experiments on human-robot interaction. Features such as ``loop functions'' in ARGoS enable some waypoint and manual control of agents, but these simulators do not have (multi-)human interaction as a central concept. More recent projects such as AirSim focus primarily on physics-modelling for ``Hardware-in-the-Loop'' simulation~\cite{airsim2017fsr} and are not intended for complex multi-robot human interaction. HARIS is a java-based Web application with low reliance on external libraries that can be easily deployed online and accessed in a web browser, making it ideal for large-scale user studies with multiple users that can access the same scenario instance simultaneously if desired. HARIS has built-in APIs for plugging in physical UAVs, allowing it to also act as a digital twin platform if desired. We use search and rescue as an exemplar of a safety-critical scenario that generalizes to other similar applications in disaster management, serving to illustrate the performance gain of human-swarm teaming in a realistic setting.

HARIS models all scenarios using a simple JSON format, meaning that all interface and algorithmic options can be customised on the fly and even the mode of operation and swarm behaviours can be changed without restarting the server. Users can also easily create their own JSON files to suit their experiment needs or use the built-in scenario editor to place targets, hazards, and agents on the realistic (google) map anywhere in the world; for example, within 5 minutes an operator could build a simulation of the actual site of a recent natural disaster for experimental and training purposes.

\section{Methodology}
To demonstrate the flexibility of HARIS, we present an example use case for human-swarm teaming in disaster relief which is to identify a number of unknown casualties in an area using a fixed number of UAVs. The methodology employed combines HARIS with a novel experimental design, to demonstrate the potential for performance gain by employing human-swarm teams in response to a real-world search and rescue scenario.

In this scenario, the swarm autonomously performs dynamic allocation using the embedded algorithms (i.e., max-sum decentralised coordination algorithm~\cite{Rogers2011,DelleFave2012}) where the swarm optimises for the total energy cost of an allocation. Human operator preferences can also be modelled as constraints used to modify the original max-sum algorithm when computing task allocation~\cite{Ramchurn2015a}. The design of HARIS is based on our previous work including a user study with 100 online participants~\cite{divband2021designing}, interviews with 6 experienced drone pilots~\cite{parnell2022trustworthy}, and a survey with 26 swarm experts from academia~\cite{naishe2022outlining}. We used the result of these studies to design and test the usability of the interface elements and functionalities such as custom interface lenses which can be switched on-the-fly to meet the operator's needs. 

\begin{figure}[ht]
  \centering
  \fbox{\includegraphics[width=0.9\linewidth]{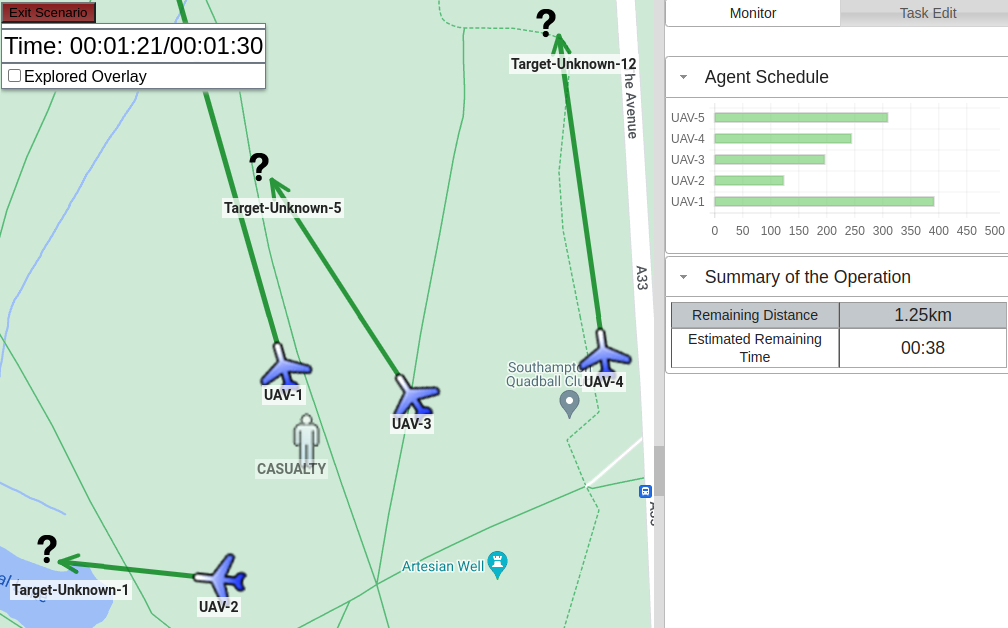}}
  \caption{The HARIS interface uses Google Maps to provide a realistic command centre for the swarm}
  \label{fig:hutsim-interface}
  \Description{The HARIS interface uses Google Maps to provide a realistic command centre for the swarm}
\end{figure}

Figure~\ref{fig:hutsim-interface} shows an experiment where four UAVs are autonomously assigned to unknown targets on a map in HARIS (these represent areas with potential casualties). The agents move towards their targets and find out whether or not they are actual casualties. The operator can monitor the views of the agents and their planned schedule on the right-hand panel, or can manually reassign them using the task view.

\begin{figure}[ht]
  \centering
  \fbox{\includegraphics[width=0.9\linewidth]{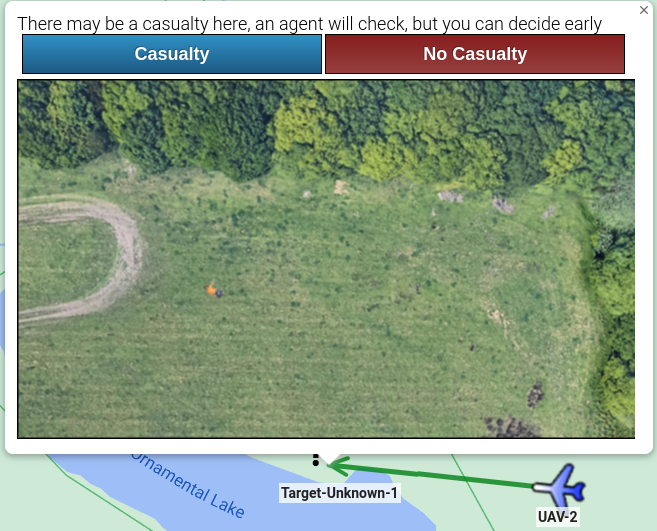}}
  \caption{Users are shown a simulated image of the target which gets clearer as the agent gets closer to the target. The user can classify the target early, saving the operation time.} 
  \label{fig:hutsim-interaction}
  \Description{An image of the Human interaction.}
\end{figure}

\section{Demonstration}

\textit{Design}: Users are presented with a Google Maps API view showing $N$ UAVs (5 by default) and $M$ unknown targets (12 by default). The swarm will autonomously use the max-sum algorithm to efficiently allocate UAVs to these targets. As the UAVs approach their targets, a popup shows a simulated camera feed of the area, which initially has a low resolution, but the resolution incrementally increases as the UAV approaches the target. 

The human user can support the swarm by viewing these images and providing an earlier classification of ``Casualty'' or ``No Casualty'' once the image becomes sufficiently clear. This allows a human to augment the swarm with their superior reasoning skills, and empower the existing strategy by reducing the number of wasted journeys, where UAVs travel to the target only to find that it is not a real casualty. Figure \ref{fig:hutsim-interaction} shows the popup with corresponding buttons for human interaction. 

\textit{Evaluation:}
Performance is assessed by a combined metric of classifications per minute $\times$ accuracy of classifications. We expect a human-swarm team to classify more targets within the time limit, leading to a higher score.  We ran the scenario using an entirely autonomous swarm (i.e, no human assistance), and provide this as a benchmark with which users can compare their performance against. This allows us to clearly demonstrate that even untrained users can empower the swarm, increasing its performance.

HARIS is easy to deploy online, and so we have made the scenario accessible, along with a video demonstrating the use case in action~\footnote{Link to \href{https://uos-hutsim.cloud/demo/aamas2023/}{HARIS Demo website}: \url{https://uos-hutsim.cloud/demo/aamas2023/}}.

\section{Conclusion}
This demonstration showcases our human-in-the-loop simulator, HARIS, by building an example use case which enables a direct comparison of untrained human users forming human-swarm teams, with fully autonomous multi-agent swarms in a safety-critical context. Further, the performance gain is measured quantitatively through a combined speed-accuracy metric for comparison between both individuals, and the autonomous swarm. HARIS is accessible online and can be used in a variety of search and rescue scenarios and will facilitate future research into the optimal strategies for human-swarm teaming.

\bibliographystyle{ACM-Reference-Format} 
\bibliography{refs}

\section*{Acknowledgements}
This work was conducted in the "Open-Source Interaction Interface for Human-Swarm Teaming" pump priming project, as part of the Trustworthy Autonomous Systems Hub, supported by the UK Engineering and Physical Sciences Research Council (EP/V00784X/1). The project also received funding from the UK Research and Innovation Centre for Doctoral Training in Machine Intelligence for Nano-electronic Devices and Systems (EP/S024298/1).  

\end{document}